\documentclass{article}

    \usepackage[preprint, nonatbib]{neurips_2020}

\usepackage[utf8]{inputenc} 
\usepackage[T1]{fontenc}    
\usepackage{hyperref}       
\usepackage{url}            
\usepackage{booktabs}       
\usepackage{amsfonts}       
\usepackage{nicefrac}       
\usepackage{microtype}      

\usepackage{graphicx}
\usepackage{amsmath}

\title{Learning Diverse Representations for Fast Adaptation to Distribution Shift}

\author{%
  Daniel Pace, Alessandra Russo, Murray Shanahan \\
  Department of Computing\\
  Imperial College London\\
  \texttt{d.pace18, a.russo, m.shanahan@imperial.ac.uk} \\
}

\begin{document}

\maketitle

\begin{abstract}
The\ i.i.d.\ assumption is a useful idealization that underpins many successful approaches to supervised machine learning. However, its violation can lead to models that learn to exploit spurious correlations in the training data, rendering them vulnerable to adversarial interventions, undermining their reliability, and limiting their practical application. To mitigate this problem, we present a method for learning multiple models, incorporating an objective that pressures each to learn a distinct way to solve the task. We propose a notion of diversity based on minimizing the conditional total correlation of final layer representations across models given the label, which we approximate using a variational estimator and minimize using adversarial training. To demonstrate our framework’s ability to facilitate rapid adaptation to distribution shift, we train a number of simple classifiers from scratch on the frozen outputs of our models using a small amount of data from the shifted distribution. Under this evaluation protocol, our framework significantly outperforms a baseline trained using the empirical risk minimization principle.

\end{abstract}

\section{Introduction}

The Empirical Risk Minimization (ERM) principle \cite{vapnik2013nature} which underpins many contemporary machine learning models is built on the assumption that training and testing samples are drawn\ i.i.d.\ from some hypothetical distribution. It has been demonstrated that certain violations of this assumption lead to models that exploit spurious correlations in the training data and are brittle with respect to certain shifts in distribution at test time. Examples include learning to exploit image backgrounds to classify cows versus camels due to a data bias \cite{beery2018recognition}, using textural as opposed to shape information to classify objects \cite{geirhos2018imagenet}, and using signals not robust to small adversarial perturbations \cite{NIPS2019_8307}.

Implicit in work that attempts to address these phenomena is the assumption that more robust predictive signals are indeed present in the training data, even if for various reasons our current models do not have the tendency to leverage them. 

Drawing inspiration from Quality-Diversity algorithms \cite{pugh2016quality} - which seek to construct a collection of high-performing, diverse solutions to a task - in this work we aim to learn a collection of models, each incentivized to find a distinct, high-performing solution to a given supervised learning problem from a fixed training set. Informally, our motivation is that a `maximal' collection of such models would exploit robust signals present in the training data in addition to the brittle signals that current models tend to exploit. Thus, given the representations computed by such a collection, it may become possible to rapidly adapt to certain shifts in distribution at test time using relatively simple models trained on frozen representations computed from relatively small amounts of new test data.

In this work we address this problem through the following contributions:
\begin{itemize}
  \item We propose a measure of model diversity based on conditional total correlation (across models) of final layer representations given the label.
  \item We estimate this measure using a proxy variational estimator computed using samples from the conditional joint and marginal distributions of final layer representations across models. We adversarially train a collection of models to be accordingly diverse, alternating between training a variational critic to maximize the above variational estimator and minimizing a weighted sum of the classification losses across models and the variational estimator.
  \item We empirically validate this framework by training on datasets for which we can measure the existence of distinct predictive signals. We demonstrate that our framework is able to learn a collection of representations which are more amenable to rapid adaptation to new test data relative to an ERM baseline. We also compare our approach to the Invariant Risk Minimization \cite{arjovsky2019invariant} framework which leverages information from multiple training environments to identify signals robust to variations across environments, noting that our approach is able to similarly exploit such stable signals from just a single training environment given relatively small amounts of data from a new test distribution.
\end{itemize}

\begin{figure}
\centering
\includegraphics[width=0.7\linewidth]{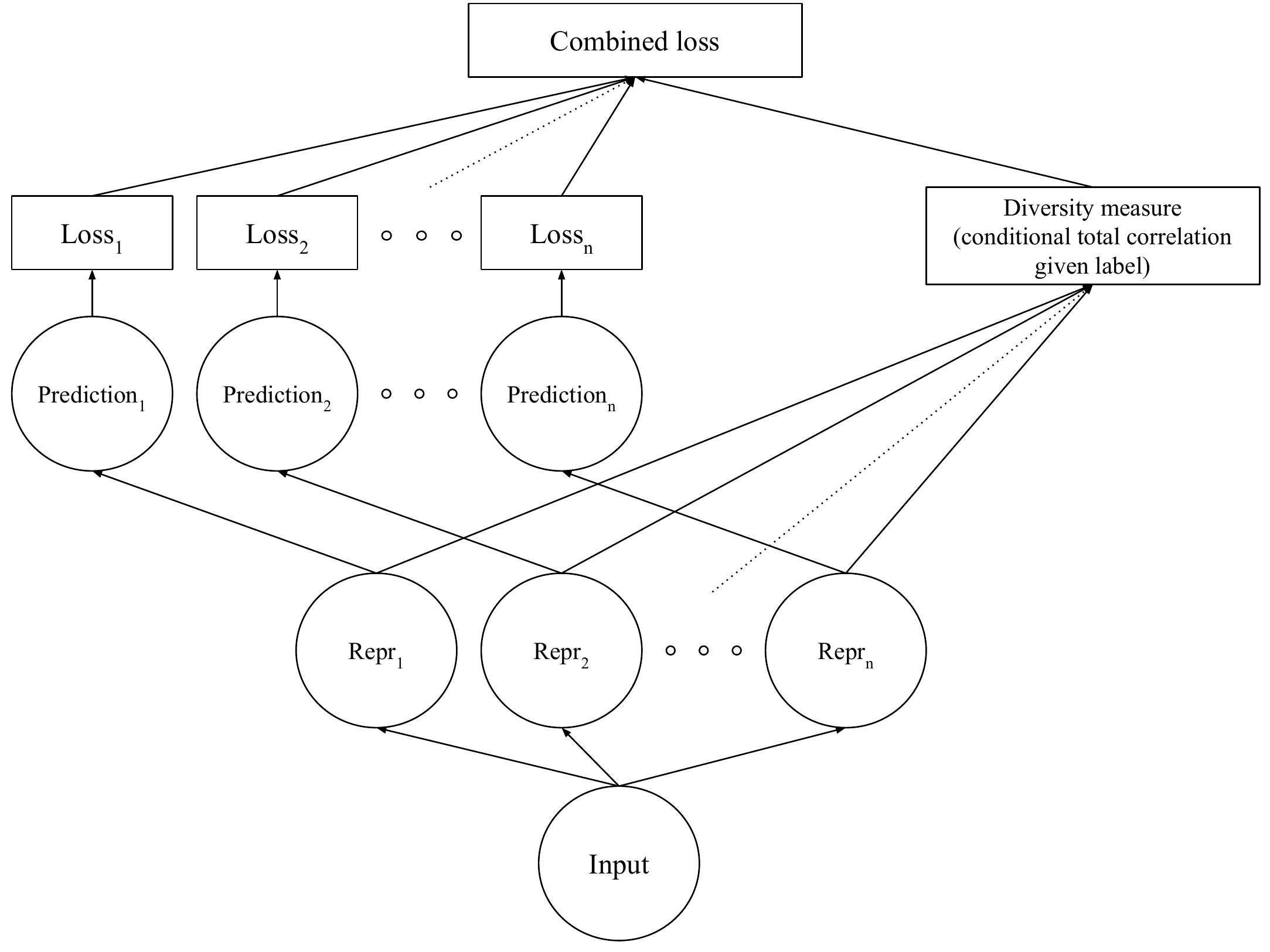}
\caption{Graphical representation of our framework. We train multiple models to individually minimize supervised loss while simultaneously minimizing conditional total correlation of final layer representations given the label.}
\end{figure}


\section{Total correlation}
Total correlation \cite{watanabe1960information} is a multivariate extension of mutual information. The total correlation for $n$ random variables $X_1, ..., X_n$ is given by:

\begin{equation}
TC(X_1, ..., X_n) := KL[p(X_1,..., X_n) \| \prod_{i=1}^{n}{p(X_i)}]
\end{equation}


We interpret this as a measure of the amount of redundancy across the $X_i$'s. A total correlation of zero corresponds to the $X_i$'s being mutually independent.

In this work, given a supervised learning problem predicting label $Y$ from $X$, we consider the conditional total correlation of a collection of vector-valued representations $h_1(X), ..., h_n(X)$ given $Y$ for differentiable functions $h_i$ , defined as:

\begin{equation}
TC(h_1(X), ..., h_n(X) | Y) :=  \mathbb{E}_{Y}[KL[p(h_1(X),..., h_n(X) | Y) \| \prod_{i=1}^{n}{p(h_i(X) | Y)}]]
\end{equation}

where $X$ is drawn from the conditional distribution given $Y$.

\section{Method}
\label{sec:method}
Consider a collection of differentiable representation function and linear classifier pairs $\{(h_i, c_i)\}_{i \in \{1,...,n\}}$ with parameters $\theta \in \mathbb{R}^k$, such that for a classification problem predicting $Y$ from $X$, we compute logits $c_i \circ h_i(X)$.

We target the following optimization problem: 

\begin{equation}
\label{opt1}
\begin{aligned}
&\min_{\theta}&&\sum_{i=1}^{n} \mathbb{E}_{X, Y}[l(c_i \circ h_i(X), Y)] \\
&\text{subject to}&&TC(h_1(X), ..., h_n(X) | Y) < \epsilon \\
\end{aligned}
\end{equation}

for some $\epsilon >0$, where $l$ is some loss function.

That is, we seek to minimize the risk for each model while keeping the conditional total correlation of final layer representations given the label below some threshold.

In practice, we attempt to solve the following unconstrained problem in place of (\ref{opt1}):

\begin{equation}
\label{opt2}
\begin{aligned}
&\min_{\theta}&&\sum_{i=1}^{n} \mathbb{E}_{X, Y}[l(c_i \circ h_i(X), Y)] + \beta  TC(h_1(X), ..., h_n(X) | Y) \\
\end{aligned}
\end{equation}

where $\beta > 0$ is a hyperparameter and we estimate the risk using the empirical risk (i.e. sample mean).

\subsection{Motivation}
\label{sub:motivation}

We discuss the following informal example to motivate our choice of conditional total correlation as a measure of model diversity:

Consider the example of classifying images of cows versus camels, where due to a sampling bias resulting in each animal tending to be located in a specific environment in the training set, the image background is a strong predictor on the training set (in addition to any signal relating to the animal). Suppose further that we have access to two hypothetical representation functions, each computing sufficient statistics for hypothetical mutually exclusive `background' and `animal' signals respectively. We hypothesize that the total correlation of these two representations may be smaller than for other pairs of representations which are not mutually exclusive (while simultaneously being sufficient to predict the label). This is based on the speculative presumption that having access to, for example, `part' of the background signal reduces our uncertainty about the `complete' background signal more than having access to `part' of the animal signal, and vice versa. Consequently, if one representation was a sufficient statistic for the animal signal as well as part of the background signal, whereas the other was a sufficient statistic for the background signal alone, knowing the former reduces uncertainty in the latter more than in the mutually exclusive case, resulting in a larger total correlation. Thus, even though hypothetical binary variables respectively representing the background and animal are strongly correlated (and hence have large total correlation), there may exist mutually exclusive representations sufficient for solving the supervised learning task while also having small total correlation. This would justify minimizing total correlation as a suitable diversity objective.

We further justify the choice of conditional total correlation in place of the unconditional version by appealing to the following argument: if we consider the case where the collection consists of just two models, total correlation is equivalent to mutual information. Minimizing mutual information between the two representations is equivalent to enforcing independence. However, it is not possible for the representations to be independent while requiring that both are sufficient to accurately predict the label. On the other hand, provided that the representations are not dependent due to some common cause, they will be conditionally independent given the label. Thus, minimizing the conditional total correlation should provide a more informative learning signal with respect to the problem we aim to solve.

In our experiments, we consider tasks that test our framework's robustness to the above assumptions.

\subsection{Total correlation estimation}
\label{sub:total_correlation_estimation}

In initial experiments with the two-model case (i.e. where total correlation is equivalent to mutual information), the InfoNCE objective of \cite{oord2018representation}, shown to compute a lower bound of the mutual information \cite{poole2019variational}, proved effective in computing useful gradient estimates for the latter term in (\ref{opt2}). Consequently, we choose to construct a proxy estimator for the total correlation based on InfoNCE.

The InfoNCE objective for random variables $X$ and $Y$ is given by:

\begin{equation}
\label{infonce}
I_{\text{NCE}} := \frac{1}{K} \sum_{i=1}^{K}{ \log \frac{e^{f(x_i, y_i)}}{\frac{1}{K} \sum_{j=1}^K  e^{f(x_i, y_j)}}}
\end{equation}

where $(x_i, y_i)$ are sampled from the joint distribution of $X$ and $Y$, $K$ is the batch size, and $f$ is a differentiable variational critic computing unnormalized scores. The parameters of $f$ are optimized to maximize $I_{\text{NCE}}$.

By analogy, our proposed estimator $\widehat{TC}(X_1, ..., X_N)$ is given by:

\begin{equation}
\label{tchat}
\widehat{TC}(X_1, ..., X_N) := \frac{1}{K} \sum_{i=1}^{K}{ \log \frac{e^{f(x_{i1}, ..., x_{in})}}{\frac{1}{M} \sum_{j=1}^M  e^{f(x_{\pi_{1,j}1}, ..., x_{\pi_{n,j}n})}}}
\end{equation}

where $(x_{i1}, ..., x_{in})$ are sampled from the joint distribution of $X_1, ..., X_n$, $K$ is the batch size, $\pi_{k,j} \sim \text{Uniform}(\{1, ... K\})$ is a random index of the batch dimension, $M$ is a hyperparameter, and $f$ is a differentiable variational critic computing unnormalized scores. The permuted $\pi_{k,j}$ indices are intended to yield $(x_{\pi_{1,j}1}, ..., x_{\pi_{n,j}n})$ samples drawn approximately from the marginals. Similarly we optimize the parameters of $f$ to maximize $\widehat{TC}(X_1, ..., X_N)$.

We interpret (\ref{tchat}) as a ratio of scores between samples from the joint distribution of the $X_i$'s and samples approximately drawn from the marginals, such that a solution to the variational objective resulting in a large ratio corresponds to being able to easily discriminate between samples from the joint and marginals (i.e. a proxy for large total correlation) and conversely when the ratio is small.

We compute the conditional case as follows:

\begin{equation}
\label{tchat_cond}
\widehat{TC}(X_1, ..., X_N | Y) := \mathbb{E}_{Y} \left[ \frac{1}{K} \sum_{i=1}^{K}{ \log \frac{e^{f(x_{i1}, ..., x_{in})}}{\frac{1}{M} \sum_{j=1}^M  e^{f(x_{\pi_{1,j}1}, ..., x_{\pi_{n,j}n})}}}              \right]
\end{equation}

where $(x_{i1}, ..., x_{in})$ are sampled from the conditional joint distribution of $X_1, ..., X_n$ given $Y$.

\subsection{Adversarial training objective}
\label{sub:minimax_objective}

We replace the latter term in (\ref{opt2}) with our variational estimator to obtain:

\begin{equation}
\label{opt_final}
\min_{\theta}\sum_{i=1}^{n} \mathbb{E}_{X, Y}[l(c_i \circ h_i(X), Y)] + \beta  \max_{\phi}\widehat{TC}(h_1(X), ..., h_n(X) | Y)
\end{equation}

where $\phi$ are the parameters of the variational critic $f$, and $\beta > 0$ is a hyperparameter.

In practice, we use gradient-based methods to approximately optimize (\ref{opt_final}) by alternating one step of optimizing $\phi$ to maximize the variational estimator, and one step of optimizing $\theta$ to minimize the sum of all terms.

\section{Experiments}

The main hypothesis that we wish to test is that our method yields a collection of models that exploit distinct predictive signals in the training data. Our strategy for testing this hypothesis is to obtain samples from a new test distribution in which only one of the predictive signals available in the training set is present, and measure the extent to which our collection of models is amenable to adaptation to this shift in distribution relative to baseline models trained with the ERM principle and the IRM method \cite{arjovsky2019invariant}.\footnote{The comparison with IRM is not like-for-like as IRM requires access to extra information relative to our approach, however we include results for contrast.}

In order to do this, we require a dataset in which such distinct signals exist and can be identified and manipulated. To that end, we consider the Colored MNIST (C-MNIST) task of \cite{arjovsky2019invariant}. This task consists of classifying MNIST digits into one of two classes (0-4 or 5-9). However, the true label is flipped with probability 0.25 and an additional binary signal (colour) is constructed from this corrupted label such that it is more strongly correlated with the corrupted label than the digit. As demonstrated in \cite{arjovsky2019invariant}, a model trained with ERM will tend to exploit the colour information. In \cite{arjovsky2019invariant}, the goal was to demonstrate an ability to learn to exploit the digit information given the task of identifying a stable signal across environments with varying colour-label correlation. To test our framework, we assume no such access to multiple environments, and collapse the two training environments in the C-MNIST benchmark task of \cite{arjovsky2019invariant} into one training set.\footnote{The two environments are constructed such that the colour signal is computed by flipping the corrupted label with probability 0.1 and 0.2 respectively. Thus, collapsing the two environments into one means that we can at best achieve an accuracy of 0.85 on the training set by exploiting the colour signal, versus the maximal accuracy of 0.75 by exploiting the digit.} We compare our approach to an ERM baseline trained on the same single training set, as well as the Invariant Risk Minimization (IRM) approach of \cite{arjovsky2019invariant}, which requires knowledge of the environments. The results for this task are collected in Table \ref{tab:cmnist}.

We also consider a modified dataset which introduces an obstacle to extracting conditionally independent digit and colour signals. Specifically, we use functions of a new binary `common cause' variable to both rotate the digits and perturb the colour signal, such that minimizing total correlation will depend on an ability to discard the common information. This is to test our framework's ability to still extract the predictive signals in the presence of such an obstacle to minimizing conditional total correlation. We refer to this dataset as Rotated Colored MNIST (RC-MNIST). The results for this task are collected in Table \ref{tab:rcmnist}.

We consider a third variation in which more than two predictive signals are present in the training data. Specifically, we construct another binary `colour' signal from the corrupted label and add it to the input such that new signal can be exploited on the training set for a maximal accuracy of 0.75. We refer to this dataset as Two-Colored MNIST (TC-MNIST). The results for this task are collected in Table \ref{tab:tcmnist}.


\subsection{Evaluation}
\label{sub:evaluation}

\begin{table}
\caption{Results for the C-MNIST task. Accuracy columns correspond to the different evaluation protocols. Test data is drawn from a distribution in which the digit information is the only predictive signal.}
\label{tab:cmnist}
\centering
\resizebox{0.7\textwidth}{!}{%
\begin{tabular}{lrrrr}
\toprule
\textbf{Method}      & \multicolumn{1}{c}{\textbf{n models}} & \multicolumn{3}{c}{\textbf{Test accuracy (s.d.)}}                                    \\
                     & \multicolumn{1}{c}{}                  & \multicolumn{3}{c}{(digit signal only)}                                                  \\ \hline
                     & \multicolumn{1}{c}{}                  & \multicolumn{1}{c}{Linear} & \multicolumn{1}{c}{Ensemble} & \multicolumn{1}{c}{Best} \\
Optimal              & \multicolumn{1}{r}{-}                 & 75                         & 75                           & 75 \\
IRM (requires env. info.)                  & \multicolumn{1}{r}{-}                 & \multicolumn{1}{r}{69.7 (1.1)} & \multicolumn{1}{r}{-} & \multicolumn{1}{r}{68.6 (1.2)} \\
Ours (conditional TC)   & 2                                     & \textbf{69.8 (0.4)}                 & \textbf{69.8 (0.5)}                   & \textbf{69.8 (0.4)}               \\
                     & 3                                     & 66.9 (0.9)                 & 68.0 (0.8)                   & 66.3 (1.8)               \\
                     & 5                                     & 65.2 (0.8)                 & 65.2 (1.8)                   & 61.8 (2.2)               \\
Ours (unconditional TC) & 2                                     & 66.7 (1.0)                 & 67.3 (1.1)                   & 66.6 (0.9)               \\
                     & 3                                     & 64.3 (0.8)                 & 67.3 (1.1)                   & 63.8 (1.9)               \\
                     & 5                                     & 61.1 (1.5)                 & 61.6 (2.0)                   & 58.9 (1.4)               \\
ERM                  & \multicolumn{1}{r}{-}                 & 56.8 (2.6)                 & -                   & 50.3 (0.7)               \\
\bottomrule
\\                      
\end{tabular}%
}

\caption{Results for the RC-MNIST task. Accuracy columns correspond to the different evaluation protocols. Test data is drawn from a distribution in which the digit information is the only predictive signal.}
\label{tab:rcmnist}
\centering
\resizebox{0.7\textwidth}{!}{%
\begin{tabular}{lrlll}
\toprule
\textbf{Method}      & \multicolumn{1}{c}{\textbf{n models}} & \multicolumn{3}{c}{\textbf{Test accuracy (s.d.)}}                                                \\
                     & \multicolumn{1}{c}{}                  & \multicolumn{3}{c}{(digit signal only)}                                                              \\ \hline
                     & \multicolumn{1}{c}{}                  & \multicolumn{1}{c}{Linear}     & \multicolumn{1}{c}{Ensemble}   & \multicolumn{1}{c}{Best}       \\
Optimal              & \multicolumn{1}{r}{-}                 & \multicolumn{1}{r}{75}         & \multicolumn{1}{r}{75}         & \multicolumn{1}{r}{75}        \\
IRM (requires env. info.)                  & \multicolumn{1}{r}{-}                 & \multicolumn{1}{r}{70.2 (0.9)} & \multicolumn{1}{r}{-} & \multicolumn{1}{r}{70.3 (0.5)} \\
Ours (conditional TC)   & 2                                     & \textbf{71.3 (0.7)}                     & \textbf{71.4 (0.6)}                     & \textbf{71.4 (0.8)}                     \\
                     & 3                                     & 66.3 (2.9)                     & 67.7 (1.7)                     & 66.8 (2.6)                     \\
                     & 5                                     & 61.4 (1.0)                     & 56.8 (2.1)                     & 57.9 (1.1)                     \\
Ours (unconditional TC) & 2                                     & 64.0 (2.6)                     & 65.5 (1.1)                     & 65.3 (1.3)                     \\
                     & 3                                     & 59.1 (1.7)                     & 55.8 (1.8)                     & 57.3 (1.8)                     \\
                     & 5                                     & 60.8 (1.2)                     & 55.2 (2.0)                     & 56.1 (0.8)                     \\
ERM                  & \multicolumn{1}{r}{-}                 & \multicolumn{1}{r}{61.6 (7.1)} & \multicolumn{1}{r}{-} & \multicolumn{1}{r}{50.4 (0.4)} \\
\bottomrule
\\
\end{tabular}%
}

\caption{Results for the TC-MNIST task. Accuracy columns correspond to the different evaluation protocols. We consider two test distributions in which the digit and new colour signals are respectively the only predictive signals present.}
\label{tab:tcmnist}
\centering
\resizebox{0.9\textwidth}{!}{%
\begin{tabular}{lccllcll}
\toprule
\textbf{Method}       & \textbf{N models}     & \multicolumn{3}{c}{\textbf{Test accuracy (s.e.)}}                              & \multicolumn{3}{c}{\textbf{Test accuracy (s.e.)}}                              \\
                      &                       & \multicolumn{3}{c}{(digit signal only)}                                        & \multicolumn{3}{c}{(second colour signal only)}                                 \\ \hline
                      &                       & Linear               & \multicolumn{1}{c}{Ensemble} & \multicolumn{1}{c}{Best} & Linear               & \multicolumn{1}{c}{Ensemble} & \multicolumn{1}{c}{Best} \\
Optimal & \multicolumn{1}{r}{-} & \multicolumn{1}{r}{75} & \multicolumn{1}{r}{75}         & \multicolumn{1}{r}{75}     & \multicolumn{1}{r}{75} & \multicolumn{1}{r}{75}         & \multicolumn{1}{r}{75}     \\
IRM (requires env. info.)                   & \multicolumn{1}{r}{-} & \multicolumn{1}{r}{67.5 (1.3)} & \multicolumn{1}{r}{-}                            & \multicolumn{1}{r}{62.8 (0.8)}                          & \multicolumn{1}{r}{75.1 (0.4)} & \multicolumn{1}{r}{-}                            & \multicolumn{1}{r}{56.3 (0.9)} \\
Ours (conditional TC) & \multicolumn{1}{r}{3} & \multicolumn{1}{r}{65.5 (0.6)} & \multicolumn{1}{r}{65.4 (0.9)}         & \multicolumn{1}{r}{65.4 (0.9)}     & \multicolumn{1}{r}{74.9 (0.0)} & \multicolumn{1}{r}{74.9 (0.0)}         & \multicolumn{1}{r}{74.9 (0.0)}     \\
ERM                   & \multicolumn{1}{r}{-} & \multicolumn{1}{r}{54.7 (3.7)} & \multicolumn{1}{r}{-}                            & \multicolumn{1}{r}{56.3 (0.5)}                          & \multicolumn{1}{r}{74.9 (0.0)} & \multicolumn{1}{r}{-}                            & \multicolumn{1}{r}{55.8 (8.5)}                        \\
\bottomrule
\end{tabular}%
}
\end{table}

We evaluate our approach by constructing test sets in which only one of the predictive signals in the training data is present. In order to test the extent to which our models are able to learn representations that leverage all present signals, in each case we train simple models on data drawn from these new distributions and record the resulting accuracy. We do the same for the representations learned by the baselines. We consider the following three such simple models, trained on the new data:
\begin{itemize}
  \item \textbf{Best}: choose the single best performing model from the collection, and use its frozen predictions ($0$ parameters)
  \item \textbf{Ensemble}: train a logistic regression model on the frozen predictions output by each model ($n$ parameters for $n$ models)
  \item \textbf{Linear}: train a logistic regression model (with $L^2$ regularization) on the concatenated frozen final layer representations from each model ($n \times D_\text{repr}$ parameters for $n$ models and representation dimension $D_\text{repr}$)
\end{itemize}

For the C-MNIST and RC-MNIST tasks, we present results for varying numbers of models in the collection to investigate our framework's ability to handle excess model capacity. We also compute results using an unconditional total correlation estimator to contrast with our proposed choice of conditional estimator. For the TC-MNIST task we report results for the best-performing model. 

We tune hyperparameters using a validation set drawn from the new test distribution and report results on a hidden test set corresponding to maximal validation scores achieved. We tune hyperparameters for IRM using the approach used to produce the results in \cite{arjovsky2019invariant}. For brevity, we exclude results for the test case where colour is the only predictive signal in the test set. It suffices to say that, as expected, a well-tuned model in every experimental condition is able to learn to exploit the colour information within a few optimization steps.

Results for each experimental condition are averaged over 10 runs, for which we report the standard deviation.

\subsection{Implementation details}
\label{sub:implementation_details}
In the notation of Section \ref{sec:method}, we use multilayer perceptrons (MLP) as representation functions $h_i$ with two hidden layers of size 128 and 64 respectively, 32 output units, and ReLu activations. We use the cross-entropy as a classification loss. Our ERM baseline is equivalent to a single model from the collection of models in our framework. The variational critic $f$ is an MLP with 2 hidden layers of size 256 computing a scalar output from an input consisting of a concatentation of all representations computed by the $h_i$'s. We use Leaky ReLus with a slope of 0.2 as activations in the critic as this appeared to speed up training in experiments. We use $\beta=10$ throughout.

We use RMSProp with a learning rate of $10^{-5}$ as an optimizer for the models in the collection as well as the critic. We note that not using a momentum-based optimizer appears to be crucial for stable training, following the trend of decreasing momentum values in the GAN literature \cite{gidel2019negative}. We note that training appears to be further stabilized by normalizing representations computed by the $h_i$'s to have unit $L^2$ norm prior to being input into the critic.

We use a batch size of 256 for training both the models and the critic, as well as $M=64$ in Equations (\ref{tchat}) and (\ref{tchat_cond}). To train the simple classifiers, we use $500$ training and validation observations from the novel test distribution, the latter used to tune the hyperparameters (if any) of the newly trained classifiers. We report test accuracies computed for the remaining $9000$ test observations. We train for 250 epochs and report test accuracies corresponding to maximal validation accuracies. We make the following exceptions for the TC-MNIST task: we train for 500 epochs and alternate between 5 steps of training the critic and 1 step of training the collection of models. This would appear to be necessary for obtaining useful total correlation gradients given the relatively more challenging variational problem that needs to be solved in this case.

\subsection{Interpretation of results}
\label{sub:interpretation_of_results}
For the C-MNIST task (Table \ref{tab:cmnist}), our approach outperforms the ERM baseline across all experimental conditions and evaluation protocols, and is competitive with IRM while also producing lower variance results. The best results are obtained using the conditional total correlation estimator and 2 models in the collection. This is not surprising given that there are exactly 2 distinct predictive signals in the training data that are conditionally independent given the label. We note that performance does appear to degrade as the number of models increases. We speculate that this is due to a combination of the variational objective becoming more challenging as the number of models increases as well as perhaps some difficulty in managing excess model capacity. Thus, we conservatively treat the number of models in the collection as another model hyperparameter. We note that the conditional TC estimator outperforms the unconditional one across all tested conditions.

For the RC-MNIST task (Table \ref{tab:rcmnist}) we see a similar pattern of results. This suggests that the framework is either successful in discarding the common information that is an obstacle to minimizing conditional total correlation, or in any case is able to produces accurate predictions on the new test set in spite of it, as desired.

The results for the `best' evaluation protocol are the clearest indication that our framework has successfully isolated different predictive signals in distinct models. The fact that the colour signal has been removed from the test data but a single frozen model from the collection is still able to produce good predictions is evidence that it is exploiting the digit information which is the only remaining predictive signal.

For the TC-MNIST task (Table \ref{tab:tcmnist}) the results indicate that our best-performing model is able to isolate all 3 predictive signals in this task into different models. Interestingly, the ERM baseline learns representations sufficient to predict the second colour signal, as indicated by the `linear' evaluation results, even though the model is primarily exploiting the primary colour signal.

\section{Related work}

Our work shares characteristics with Quality-Diversity (QD) algorithms \cite{pugh2016quality} from the evolutionary computation literature, which also attempt to find diverse, high-performing solutions to a task. This is typically done by partitioning the search space using a predefined behaviour descriptor and finding high-performing solutions within each behavioural niche \cite{mouret2015illuminating,cully2017quality}. In contrast, we do not explicitly partition the search space but instead enforce diversity through an information theoretic objective. Indeed, the central challenge in directly applying existing QD algorithms to the contexts considered in this work is the difficulty in defining an appropriate behaviour descriptor. An interesting direction for future work would be to investigate the possibility of learning suitable behaviour descriptors from data, enabling the use of existing QD algorithms for supervised learning problems. In the context of a control problem, \cite{cully2019autonomous} uses the latent code of a learned generative model as a behaviour descriptor. However, it is not clear whether partitioning the search space on the basis of such a behaviour descriptor will enforce a notion of model diversity that is effective in addressing the problems we target in this work.

Ensemble-based methods \cite{polikar2006ensemble,rokach2010ensemble} train collections of models using various techniques (e.g. subsampling of observations and features \cite{breiman2001random}, boosting \cite{friedman2001greedy}, penalizing the correlation of model predictions \cite{liu1999ensemble}) to enforce some notion of model diversity such that aggregate predictions outperform single model predictions. It is unclear whether these heuristics are able to enforce a notion of diversity suitable for the problem we target in this work, nor whether they are appropriate for high-dimensional, unstructured data. We note that our `ensemble' evaluation protocol is related to the `stacking' method from this literature \cite{wolpert1992stacked}. 

A number of works \cite{kim2018disentangling,chen2018isolating,poole2019variational,gao2019auto} use total correlation based objectives to enforce factorized representations in generative models. Similar to our approach, FactorVAE \cite{kim2018disentangling} uses adversarial training to minimize an estimator of total correlation. In contrast to these approaches, we target total correlation for a collection of high-dimensional vector-valued representations while also leveraging label information. A possibility for future work is to impose probabilistic structure on the representations in our proposed method that may enable the use of the upper bound on total correlation presented in \cite{poole2019variational}, perhaps alleviating the need for adversarial training.

Independent component analysis (ICA) \cite{comon1994independent} addresses the problem of separating a signal into distinct components using an independence based criterion. This is done in an unsupervised setting and making different assumptions than we do. \cite{brakel2017learning} use an adversarial training method similar to ours in the context of non-linear ICA.

Other works \cite{kwok2012priors,mariet2015diversity,pang2019improving,kwok2012priors} use determinantal point processes \cite{kulesza2012determinantal} to compute diversity measures. This would also be interesting to investigate in the context of the problems addressed in this paper.

This work ultimately shares the same motivation with other attempts to yield models that generalize better under certain violations of the ERM assumptions. Invariant Risk Minimization \cite{arjovsky2019invariant} assumes access to data from multiple environments and attempts to find a representation such that the same classifier is simultaneously optimal in all environments (as do subsequent works addressing the same / similar problem \cite{ahuja2020invariant,krueger2020odd}). The problem of causal discovery \cite{peters2017elements} consists in trying to identify underlying causal structure from data such that the tools of causal inference \cite{pearl2009causality,peters2017elements} can then be used to compute predictions that are valid under certain shifts in distribution. While our work does not attempt to tackle the challenging task of identifying which signals in observational data are related to a causal explanation of interest, it would be interesting to investigate whether our framework could facilitate some aspect of causal discovery and/or the identification of invariant signals.

\section{Conclusion}
We have introduced a method for training a collection of high-performing supervised learning models each incentivized to learn a distinct way to solve the task. We have empirically demonstrated how minimizing our proposed measure of model diversity in addition to the empirical risk suffices to yield a collection of models each exploiting distinct predictive signals present in benchmark tasks.

We have focused on evaluating our approach on datasets which we have sufficient control over to enable the testing of basic hypotheses about our proposed framework. To the best of our knowledge, there are currently no other more complex, natural datasets for us to evaluate our approach on. Furthermore, as in \cite{arjovsky2019invariant}, we require access to evaluation data from a different distribution than the training data in order to tune hyperparameters. Rather than seeing this as a disadvantage of our approach, we see it as further evidence that confronting the problem of generalization in a data-driven way may call for the development of new kinds of datasets and benchmarks that are better suited to testing robustness to violations of the\ i.i.d.\ assumption on which the empirical risk minimization principle is founded.

\section*{Acknowledgements}
We thank Antoine Cully for helpful discussions.



\bibliography{bibliography}
\bibliographystyle{abbrv}

\end{document}